\documentclass{article}

\usepackage{arxiv}

\usepackage[utf8]{inputenc}
\usepackage[T1]{fontenc}
\usepackage{lmodern}
\usepackage[british]{babel}
\usepackage{amsmath}
\usepackage{amssymb}
\usepackage{amsfonts}
\usepackage{booktabs}
\usepackage{graphicx}
\usepackage{natbib}
\usepackage[hidelinks]{hyperref}
\usepackage{url}
\usepackage{microtype}

\title{Trusting sovereign language models as scientific instruments: evidence from Portugal's AMALIA}
\renewcommand{\shorttitle}{Trusting sovereign language models}

\author{
  Manuel Pita \\
  Artificial Intelligence, Social Interaction and Complexity Laboratory \\
  CICANT, Universidade Lus\'ofona \\
  \texttt{manuel.pita@ulusofona.pt}
}

\date{July 2026}

\begin{document}
\maketitle

\begin{abstract}
National language models are becoming publicly funded epistemic infrastructure. Public ownership, linguistic specialization, and open weights create a presumption of trustworthiness. Such an instrument, built by and for a language community, looks like the natural choice for measuring what that community says and values. Whether such a model validly measures anything is untested at release. The evaluation of LLMs as measurement instruments is typically task-specific and stops at agreement with human coders. Agreement cannot distinguish an LLM instrument that measures a construct from one that reaches matching codes through surface correlates. We audit the presumption on a favourable case: AMALIA, Portugal's publicly funded 9B model, coding the moral foundation of authority in European Portuguese. The \textit{recovery gap} operationalizes the audit: decompose the codebook into its theory-defined clauses, recombine them through the theory's explicit rule, and measure how much of the original prompt's performance the stated theory reproduces. In a pre-registered, out-of-sample study on a transcreated (English to European Portuguese) corpus, AMALIA agrees with trained coders within six points of open models eight to thirteen times its size. Yet, the recovery gap shows that only about half of coding performance on authority can be attributed to the theory. A larger multilingual LLM closes the recovery gap on the same corpus, suggesting the shortfall lies in the annotator model, not the corpus or its translation. Sovereignty earns operational and performance trust; epistemic trust requires calibration---and the audit method is inexpensive, and portable across models, languages and tasks.
\end{abstract}


\section{Introduction}\label{sec:intro}
On 1 July 2026, Portugal released AMALIA, a nine-billion-parameter language model\footnote{\url{https://huggingface.co/amalia-llm/AMALIA-9B-0626-DPO}}, built with public funds to serve European Portuguese \citep{simplicio:2026a}.
The release joins a growing family of national and sovereign models, built so that a public institution can hold, inspect and operate a model without depending on the small set of private organizations that currently concentrate access to these systems \citep{pipatanakul:2026}.
The motivation is both empirical and sociopolitical.
Empirically, the dominant LLMs were trained mostly for use in English \citep{rao:2025}.
Sociopolitically, a nation's capacity to hold and audit these systems has become part of a wider agenda of digital sovereignty, pursued at the institutional level \citep{fitsilis:2026}.

None of these new local LLMs claims general superiority over large-scale commercial models.
The claim is, rather, superior performance on locally relevant tasks, in the language of the community it sets out to serve.
The performance evidence AMALIA was released with follows the family pattern.
Its most distinctive Portuguese-specific benchmarks were built by the very team that created the model, from open-ended task results such as tests of linguistic competence and of fidelity to the European variant, scored by another language model in the role of judge \citep{simplicio:2026a}.

The pattern matters because public ownership changes how a model is received.
A system built with national funds, released with open weights and documented for the community it serves arrives with a form of institutional credit no commercial API enjoys: the institutions that depend on it can hold, inspect and audit it \citep{pipatanakul:2026}.
Trust in the instrument is, in part, trust in its governance.

That credit invites a quiet substitution.
A model trained to represent a language variety well is easily taken for a competent instrument for research in that variety, as if fidelity to the language certified fidelity to whatever researchers measure through it.
These are different properties.
Representing European Portuguese is a property of a model's text; measuring a theoretical construct in European-Portuguese text is a property of an instrument's inferences, and it must be established separately.

The distinction is institutional as much as technical.
A publicly funded model that becomes part of a country's knowledge-production infrastructure inherits the obligations of infrastructure: measurements taken with it will inform research about, and eventually policy for, the community that paid for it.
On this view, application-specific validity is not a technical refinement but part of what responsible stewardship of sovereign AI requires.
The presumption of trustworthiness that sovereignty creates is precisely what a validity audit puts to the test.

One of the most important locally relevant tasks lies in the domain of text-as-data.
In the social and behavioural sciences, AMALIA could become a critical resource for annotating theoretical constructs in text.
This kind of annotation allows, for example, the study of collective phenomena from what people say on social media, group debates in schools, interviews and open-ended surveys, or in the study of parliamentary activity, among others.
By a theoretical construct we mean an analytical category that is not directly observable via surface features, and so corresponds not to a word, an expression or a topic, but to a property inferred from text in the light of a theory.
Morality, for example, can be represented by constructs such as \textit{authority}, \textit{harm}, \textit{care} or \textit{loyalty}.
But recognizing the presence of authority in a passage is not the same as finding the word `authority' or words related to it.
The difficulty is that recognizing authority requires identifying a relation between who must obey, who holds authority and whether that relation is affirmed, contested or violated.
The concept is formalized in §\ref{sec:background}.

The immediate point is that, for one of the fundamental uses the social sciences will make of these models, namely the annotation of constructs, the record available at AMALIA's release does not have anything to say yet \citep{simplicio:2026a}.
This silence is not a limitation of AMALIA.
Evaluating LLMs as construct annotators is complex, and has so far centred mainly on agreement with human annotators \citep[see, for example,][]{abdurahman2024perils, rathje2024gpt, gilardi2023chatgpt}, rather than on the model's validity as a measurement instrument.

This practice inherits the norm of content analysis, by which the quality of annotation is settled by agreement among annotators \citep{krippendorff2018content}.
But in the case of LLMs as coders, agreement only tells us the instrument is reliable---not that it is \textit{valid}.
When the annotator is an LLM, a fundamental question is whether the instrument measures the construct through the inferences specified by the underlying theory, or reaches codes that agree with human annotators by other routes correlated with the construct.

Answering this question matters because LLMs match human annotators on many annotation tasks.
That agreement motivates the adoption of LLMs as far cheaper instruments, operating at scale to process large volumes of data.
Where the correlate and the construct diverge, however, LLMs fail invisibly because no agreement metric tells a genuine hit apart from a gross error.
An LLM can thus produce the right codes for the wrong reason, without measuring what the construct's theory specifies.
The starting point with LLMs is then a \textit{theoretically naive instrument} \citep[][]{pita:2026a}.

For a national model such as AMALIA, agreement in the community's own language will tend to be read as the success of its agenda and programme as a whole.
Whether that agreement measures the construct or merely reproduces it through surface correlates calls for an operational test.
This work centres on a construct from moral foundations theory, \textit{authority/subversion}, annotated by means of \textit{grain calibration} \citep{pita:2026a}.

The study has two objectives.
The first is to evaluate how well AMALIA can function as a coding instrument.
The second is to present what we believe is the first validity study of an LLM-based instrument for measuring theoretical constructs within a national model and the first conducted in European Portuguese.

We replicate, in European Portuguese and with AMALIA, a study originally conducted in English, and compare the results.
The analysis is guided by four questions:

\begin{enumerate}
\item Can AMALIA annotate the construct as the theory defines it, or does it arrive at the correct codes by alternative routes, such as surface correlates?
\item Does the language of the prompt instructions, Portuguese or English, applied to the same Portuguese texts, change what AMALIA measures?
\item How does AMALIA compare with the large multilingual models against which it is often positioned, on the same Portuguese corpus, and how much does each model lose relative to its performance on the English corpus?
\item Does grain calibration performed in one language transfer to another?
\end{enumerate}

The study was pre-registered before any confirmatory annotation began.\footnote{Zenodo: 10.5281/zenodo.21178967}
Hypotheses, endpoints, interpretation bands and error-analysis criteria, together with a pilot study on a balanced subcorpus of 300 texts, were fixed in advance.
All confirmatory tests are run on the remaining 448 texts that no model had seen.

The goal of §\ref{sec:background} is to introduce the relevant theory, the grain calibration method, the construct under study, and this paper's central measure, the recovery gap.
Then §\ref{sec:method} describes the method, including the transcreation that converts the 748 annotated texts of the original English study into European Portuguese, under several verification gates.
The results are presented in §\ref{sec:results}, which reports the comparative findings between AMALIA and two open-weight LLMs.
Finally, §\ref{sec:discussion} draws out the consequence for national-model programmes, and returns the question with which the country received the model.

\section{Background}\label{sec:background}
Consider a text that mentions a president amid moral outrage about another matter.
It contains the whole vocabulary of authority/subversion (order, duty, rebellion), yet whether it instantiates the construct turns on an evaluative stance towards hierarchy that the vocabulary alone cannot settle \citep{pita:2026a}.
A model keyed to that vocabulary codes every such text as authority, and still agrees with human coders wherever the authority-holders are in fact being appraised.

The cost is already on record.
For some constructs, a zero-shot prompt \citep{brown:2020} inflates false positives roughly tenfold relative to trained coders \citep{abdurahman2024perils, pita:2026a}.
If the instrument is reliably right, does the reason still matter?
It does, because agreement cannot localize the failure.
When the model's codes diverge from the human ones, the error cannot be localized finely enough to tune the instrument to the construct rather than to its correlates.

The \textit{grain calibration} method was introduced precisely to remove this opacity of the instrument \citep{pita:2026a}, and consists of five steps:

\begin{enumerate}
	\item Compute the instrument's performance with the \textit{undecomposed} prompt, designed from the codebook.
	\item Decompose the undecomposed prompt into the units the construct's theory defines (\textit{components} and, within them, \textit{clauses}), at a grain at which the LLM can reliably answer from observable surface features, as opposed to surface correlates reached in ways we cannot trace.
	\item Consolidate the annotated units into a code through an explicit \textit{integration rule}, derived from the theory.
	\item Quantify the degree to which the decomposition recovers the undecomposed prompt's performance.
	\item At this point, one can enter a loop in which the prompt is calibrated through clause adjustments, propagated to the undecomposed prompt, to reduce the performance difference between the two and improve the instrument as a whole.
\end{enumerate}

An instrument validated this way satisfies the \textit{substantive aspect of construct validity}.
That is, it provides empirical evidence that the instrument measures the construct as the theory defines it, and not through a correlate standing in for it \citep{messick1995validity}.
The recovery gap concerns this aspect alone: it does not establish criterion or predictive validity, generalization across populations and registers, or the validity of the construct itself in Portuguese.

To make the decomposition concrete, consider that most theoretical constructs specify (a) a \textit{detection} component, by which observables in a text establish that the construct's domain is at issue; and (b) a \textit{distinction} component, which separates the construct from neighbours that share observable features.
Some constructs add further components, for example an \textit{appraisal} component, which requires evidence that the text evaluates what is detected.
Components decompose into \textit{clauses}: atomic yes/no questions about single observables, or binary relations between them, cut to what the LLM's distributional competence can answer reliably \citep{pita:2026a}.
The idea of decomposing a construct is one of the foundations of measurement theory.
A construct is defined by a \textit{nomological network} of propositions about observables and about the expected relations between them, which allows the underlying theory to be tested \citep{cronbach1955construct}.

The construct at the focus of this study is \textit{authority/subversion}, which is bound up with the moral ordering of power and deference \citep{graham:2013a}.
This moral foundation spans legitimate authorities and high-status individuals, from leaders and governments to the police, the courts, employers, parents, elders, and religious and collective authorities, and it spans the virtues of leadership, obedience, duty and respect for tradition, alongside the vices of disorder, defiance of authority and resentment of hierarchy.
The codebook of the Moral Foundations Reddit Corpus \citep[MFRC,][]{trager2022moral} follows the revised six-foundation taxonomy \citep{atari:2023}.
A text is annotated positive for authority if the annotator verifies that it,

\begin{enumerate}
	\item takes a stance on obedience, deference or order, whether in favour (appealing to obedience, duty or order) or against (appealing to defiance, resistance or overthrow);
	\item appraises an authority, institution or candidate for power in its role of power, as to conduct, fitness or legitimacy to hold it, in a judgement that would make no sense about an ordinary person;
	\item or takes a position towards obedience, hierarchy, tradition or order as values (defence or rejection).
\end{enumerate}

Examples that do not express authority include texts that merely identify an entity with power without taking a stance, that voice a generic reaction with no identifiable basis in authority or order, or that merely disagree with an authority's policies or decisions without judging it as a holder of power.
It is this demand for a stance, together with the distinction between criticizing a policy and judging whoever exercises power, that makes the construct hard to annotate.
Its surface vocabulary (duty, tradition, riot, disobey) is shared with the neighbouring foundations and with non-moral political talk, and indeed with other moral-foundations constructs, among them \textit{loyalty}.

This construct's codebook can be operationalized at codeable grain.
Seven binary clauses are organized into the three components introduced above (Table~\ref{tab:clauses}).
The \textit{detection} clause ($D$) establishes that the text refers to an authority, an authority-holder or the social order; it is the construct's anchor, a deliberately broad, recall-oriented detector (rulers, police, parents, God, laws, tradition, `the system', candidates for power).
The three \textit{distinction} clauses detect the neighbouring foundations that share the vocabulary (equality, proportionality and loyalty); by design they serve only as diagnostics, so they were not used as exclusion mechanisms, and therefore fall outside the decision rule.
The three \textit{appraisal} clauses establish that the text takes an evaluative stance:
(a) towards obedience, deference or order, as a speech act, in favour or against ($A_1$); (b) on the conduct, fitness or legitimacy of whoever exercises power ($A_2$); and (c) on obedience, hierarchy, tradition or order as values ($A_3$).
Each clause is an atomic question, answered independently, that always includes a segment of textual evidence.
The explicit integration rule, informed by the theory and fixed before any work in Portuguese, combines four of these clauses into the construct decision, as set out in Logical Expression~\ref{eq:authority-rule}.

\begin{equation}
\label{eq:authority-rule}
	(D \land (A_1 \lor A_2)) \lor A_3 \Rightarrow \texttt{authority}
\end{equation}

\noindent
The rule has two paths to the moral foundation `authority'.
One runs through the anchor $D$, together with an appraisal of its function as an authority ($A_1$) or of the authority itself as an entity ($A_2$).
The other corresponds to an appraisal of order as a moral value ($A_3$).
This route dispenses with explicit identification of the entity, the anchor, because, on the theory, appraising the hierarchy of power is sufficient to conclude that the construct is expressed.
By contrast, note that detecting the anchor, $D$, is never sufficient.
Without appraisal, the anchor formalizes the construct's first negative condition, namely that the text merely names an authority without taking a stance.
The evaluation is deterministic, a Boolean expression over the binary answers, informed by the theory, with no weights, thresholds or tuning, so that nothing stands between the clause answers and the final code.

\begin{table}[htb]
  \centering
  \small
  \begin{tabular}{@{}p{2.2cm}p{2.0cm}p{5.4cm}p{4.2cm}@{}}
    \toprule
    \textbf{Clause} & \textbf{Component} & \textbf{Question answered} & \textbf{Role in the rule} \\
    \midrule
    $D$ & Detection & Authority-holder or the social order present? & Anchor; required with $A_1$ or $A_2$ \\[2pt]
    $A_1$ & Appraisal & Stance on obedience, deference or order, as a speech act? & One appraisal arm (with $D$) \\[2pt]
    $A_2$ & Appraisal & Judgement of a power-holder's conduct, fitness or legitimacy? & One appraisal arm (with $D$) \\[2pt]
    $A_3$ & Appraisal & Stance on hierarchy, tradition or order as values? & Sufficient on its own \\[2pt]
    Equality, proportionality, loyalty & Distinction & A neighbouring moral foundation instead? & Diagnostic only; not in the rule \\
    \bottomrule
  \end{tabular}
  \caption{The seven clauses, their components, and their role in the integration rule $(D \land (A_1 \lor A_2)) \lor A_3 \Rightarrow \texttt{authority}$. The three distinction clauses are diagnostic and do not enter the decision.}
  \label{tab:clauses}
\end{table}

The final undecomposed and decomposed prompts result from grain calibration over seven cycles, against the English MFRC ground truth used as the reference \citep{pita:2026a}.

The model is a black box: we see the code it returns, but not the inference that leads to it. Decomposing the construct allows us to probe the box systematically, enabling us to diagnose how the LLM likely arrives at its code.
We do this not by opening the box but by testing whether the LLM can reach the same code through the decomposed, clause-level route.
The same construct can be annotated in two ways that matter to the calibration method.
Annotated with the undecomposed prompt, the LLM receives the whole codebook in a single prompt and returns a single answer.
Annotated with the decomposed prompt and the integration rule, the LLM answers each clause independently, and the rule combines the answers to produce the code.
If the two variants of the instrument agree equally well with the ground truth, we accept that the undecomposed prompt annotates the construct according to the theory that defines it.
When the undecomposed and the decomposed prompts disagree with the ground truth on a sizeable number of examples, the likeliest explanation is that the undecomposed prompt is reaching the correct result by the wrong routes.
The \textit{recovery gap}, $\Delta$, quantifies this divergence. Formally,

\begin{equation}
	\Delta = F_1^{\mathrm{u}} - F_1^{\mathrm{d}}
\end{equation}

\noindent
where \textit{u} stands for undecomposed and \textit{d} for decomposed.  
Both $F_1$ values are computed against the same ground truth.\footnote{The pre-registered protocol refers to the undecomposed prompt as the \textit{holistic} prompt.}
A value of $\Delta \approx 0$ indicates that the decomposition fully recovers the undecomposed prompt's performance.

The larger the value, the more the undecomposed prompt's performance rides on evidence the theory does not define as valid for the construct.
Possibly, the instrument uses `shortcuts' that the decomposition exposes. The difference signals that shortcuts exist, but not which. Their identification thus requires clause-level error analysis.
Observing $\Delta \approx 0$ does not confirm that the undecomposed prompt reaches the correct codes by measuring the construct; it confirms that, through the evidence the decomposed prompt supplies, the instrument can measure the construct when specified at a grain at which the clauses can be resolved \citep{pita:2026a}.
The value of $\Delta$ can also be negative.
Such a value means that composing the codes from the clauses outperforms the undecomposed prompt's judgement.
The instrument can meet the criteria one by one, but fails at their integration as implemented in the undecomposed prompt.
The cause may be that the undecomposed prompt does not match the decomposed prompt or, if the two are equivalent, the negative value indicates that the instrument should be applied in several separate steps and integrated through an explicit rule.

What matters for replicating in Portuguese a study originally done in English---the methodological basis of this work---is that $\Delta$ is a property of the construct--model pair. The measurement instrument is neither the model alone nor the construct alone, but the construct operationalized as a prompt and executed by a model. The same theoretical specification can therefore measure the construct in one model and fail in another.
On the original English corpus, the same instrument calibrated for the `authority' construct closes the recovery divergence with the LLM GPT-OSS-120B ($\Delta = 0$), and sits at the threshold of tolerance with the LLM Llama-3.3-70B ($\Delta = 0.054$), on the same corpus.
The starting point of this study is therefore that, in English, we have an LLM instrument calibrated to measure authority in text passages. Beyond this study, the recovery gap is a portable diagnostic for any LLM used as an annotator. Where the difference is large, the model returns the right codes without the inferences the theory defines.

AMALIA-9B is Portugal's national language model.
It continues the pretraining of EuroLLM-9B \citep{martins:2025} over a mixture that adds code, long-context data and 5.8 billion tokens of curated European-Portuguese web text, distilled from the Arquivo.pt archive (\url{https://arquivo.pt}).
The context window is extended from 4K to 32K tokens \citep{simplicio:2026a}.
Post-training has two stages.
Supervised fine-tuning targets instruction following, conversational reasoning, mathematics and safety.
Its European-Portuguese portion is largely synthetic, generated with Gemma 3-27B.
Alongside it, two resources are built by hand: 200 Portuguese linguistic instructions curated by a linguistics expert, and 156 self-referential entries about the model itself.
Preference training is direct preference optimization over 478K pairs.
Candidate responses are sampled from the SFT model and ranked by the ArmoRM reward model.
Two chat variants are released, AMALIA-9B-SFT and AMALIA-9B-DPO.
This study evaluates the flagship version, 0626-DPO.

The technical report evaluates knowledge, linguistic competence, generation fidelity and safety.
Its reference pt-PT results are scored by an external language model in the role of judge.
The AMALIA LLM was not trained on annotation or classification tasks, and its technical report reports no external evaluation of the model as an annotation or classification instrument \citep{simplicio:2026a}.
The home-language promise, that a national model is the natural research instrument for its language, is therefore, at the date of release, a hypothesis.

The evidence base for LLM annotation stops one layer short of that hypothesis.
Agreement studies established broad adoption.
Overall, LLMs match or exceed crowd workers and, on some tasks, trained annotators \citep{gilardi2023chatgpt, tornberg2024chatgpt}.
For Portuguese, the existing evaluation bears on the Brazilian variant.
Sabi\'a-3 exceeds 92\% mean accuracy on binary sentiment benchmarks \citep{schuck:2025}.
The same model reaches only a mean $F_1$ of 25.5\% on argument mining \citep{pereira:2025}.
This last study is evidence that training the LLM in the language of the dataset does not guarantee an advantage over multilingual models.
The lesson extends beyond Portuguese.
In low-resource languages, benchmark standing likewise fails to transfer to annotation; in a Marathi test, even top models fall 10.2 and 14.1 accuracy points behind BERT baselines fine-tuned for the task \citep{jadhav:2024}.
This is precisely the condition AMALIA's own authors diagnose for pt-PT: under-representation in the training data and in native evaluation \citep{simplicio:2026a}.
Across this literature the test is always agreement: the codes the LLM produces are compared with human codes ($F_1$, kappa).
No evaluation of a language-specific model quantifies the capacity to annotate theoretical constructs beyond agreement, which is the focus of the study described in the sections that follow.

\section{Method}\label{sec:method}
This study compares AMALIA's performance in annotating one of the constructs in moral foundations theory, authority, with that of two other large-scale open-weights LLMs.
The Portuguese corpus is a European-Portuguese transcreation ($C_{\mathrm{pt}}$) of 748 texts of the \textit{authority} construct drawn from the MFRC ($C_{\mathrm{en}}$), built under the verification gates described below; by transcreation we mean translation that preserves referents, stance and illocutionary force while adapting idiom and register to the target variety, here Reddit-colloquial European Portuguese.\footnote{The ground-truth corpus derives from the author's deduplicated reconstruction of the MFRC. The released distribution contains duplicate (text, annotator) rows (25.1\% of rows, affecting 42.6\% of unique texts) that majority voting double-counted. Deduplication reduces authority's positive codes from 1063 to 663, a 37.6\% reduction. The MFRC authors \citep{trager2022moral} confirmed the artefact (personal communication, 18 May 2026).}

The codebook's definition in the instrument exists in two versions.
The first is the fixed English prompt, $P_{\mathrm{en}}$, with undecomposed and decomposed variants.
The second is the translated pt-PT version, $P_{\mathrm{pt}}$, which keeps components, clauses and integration formula identical.
Each instrument therefore consists of an LLM, $m$, operating a version, $\ell$, of the prompt, $P^{m}_{\ell}$, with $m \in \{A,L,G\}$ (AMALIA, Llama-3.3-70B, GPT-OSS-120B) and $\ell \in \{\mathrm{en},\mathrm{pt}\}$.

AMALIA, model $A$, is the only model tested under both prompt versions, $P^{A}_{\mathrm{en}}$ and $P^{A}_{\mathrm{pt}}$.
The reference models, Llama-3.3-70B and GPT-OSS-120B, are the two best-performing annotators from the English calibration study, and operate only the English-defined instrument $P_{\mathrm{en}}$, as $P^{L}_{\mathrm{en}}$ and $P^{G}_{\mathrm{en}}$.
All instruments annotate the Portuguese corpus, $C_{\mathrm{pt}}$.
The answer tokens always remain in English (\texttt{\{"answer": "yes"|"no"\}}) in every condition, with no effect on the nature of the study.

Each text in a corpus, $C$, produces up to eleven separate annotations per condition.
The undecomposed prompt is a single call.
The decomposed prompt uses three cumulative build-up variants for the detection, distinction and appraisal components, and seven for clause extraction, recomposed by the integration formula.
The reference points for all comparisons are the same models' results on $C_{\mathrm{en}}$.

The construction of $C_{\mathrm{pt}}$ follows four pre-registered rules.
The first makes generation blind to the codes present in the MFRC.
That is, the LLM that generates the Portuguese texts never sees the ground-truth codes, nor the contest status (unanimous or contested), so code preservation must come from preserving the text's meaning.
The second preserves meaning and stance.
Each illocutionary act, whether an order, an accusation, praise or sarcasm, keeps the same target and the same intensity.
The third keeps referents stable.
People, parties and institutions keep their names, and only idiom, slang and register are localized.
The fourth fixes the target register as European Portuguese, in Reddit's colloquial style.
An amendment corrected a defect in the third rule, which, applied to the letter, left place names untranslated.
The amended version, adopted before the main analyses, keeps referents stable but gives place names the standard Portuguese exonym and translates generic institutional descriptors.
The ground-truth codes are inherited from the source texts, and the validity of that transfer is defended by verification, not assumed; the residual risk of drift under transcreation is treated as a limitation (§\ref{sec:discussion}).

Corpus construction is sealed off from evaluation.
The evaluated LLMs, AMALIA, Llama and GPT-OSS, take no part in the transcreation of $C_{\mathrm{pt}}$, so as to rule out any kind of contamination.
The transcreation pipeline uses DeepSeek-V4-Pro to generate, Qwen3-235B to verify, and DeepSeek-V4-Pro to back-translate (sampling parameters in Appendix~A).

Every text passes a five-criterion machine gate (meaning, stance, pt-PT variety, naturalness, entity stability), regenerates with feedback on failure, and, if attempts are exhausted, goes to human adjudication whose decision outranks the gate.
Table~\ref{tab:funnel} reports the funnel; gate criteria, regeneration protocol and per-text adjudication records are given in Appendix~A.

Because the gate reads fluency, two stratified back-translation audits of sixty texts each, one per phase, were read against the English sources; each caught exactly one error the gate had passed, a catch rate of 1.7\% beyond the gate (sampling and audit protocol in Appendix~B).
The frozen corpus, $C_{\mathrm{pt}}$, contains the 748 texts of $C_{\mathrm{en}}$, with strata identical to the source: 322 unanimous negatives, 317 contested positives, 57 unanimous positives and 52 contested negatives.
The corpus contains every text annotated positive for authority in the MFRC, most of which were not a unanimous decision among the human annotators.

The English construct file is byte-for-byte identical to the final English calibration.
We do not recalibrate the instrument for AMALIA: the object under test is the transfer of a calibrated instrument, not the best attainable AMALIA-specific one.
The Portuguese version translates only natural language, and clause identities, the per-component layer structure, the integration formula and the answer tokens remain intact.
Because the translated construct is the instrument in the pt-PT condition, it underwent qualitative validation.
A native European-Portuguese analyst validated both $P_{\mathrm{pt}}$ variants, undecomposed and decomposed.

All annotation runs at temperature 0, with output constrained to JSON and a logged fallback tier for responses that resist structured parsing.
A pre-registered robustness pass repeats AMALIA's undecomposed prompt with no output constraint, measuring raw format discipline.
The full study consumed 2.35 GPU-hours, and about ninety minutes of human reading of the verifications.\footnote{The total cost of the study was EUR~30, which is about US \$35.}
There is no new human annotation cost, because the ground truth carries over from the source corpus.

The confirmatory protocol was deposited on Zenodo (DOI: 10.5281/zenodo.21178967) after the pilot and before any model had annotated the 448 new texts of the main phase.
The registration declares the pilot as the source of the hypotheses.
All confirmatory tests therefore bear only on the 448 previously unseen texts, and the full-corpus figures are reported as descriptive.

The primary endpoint is the recovery gap per instruction condition, $\Delta^{m}_{\ell}(C) = F_1^{\mathrm{u}} - F_1^{\mathrm{d}}$, where $F_1^{\mathrm{u}}$ and $F_1^{\mathrm{d}}$ are the undecomposed and decomposed prompt's performance.
Each $F_1$ is the positive-class score for the authority label, $2\,\mathrm{tp}/(2\,\mathrm{tp}+\mathrm{fp}+\mathrm{fn})$.
The interpretation bands are fixed in advance: $\Delta \geq 0.10$ open, $\Delta < 0.05$ closed, and the interval between the two indeterminate.
H1 states that $\Delta^{A}_{\mathrm{en}}(C_{\mathrm{pt}})$ and $\Delta^{A}_{\mathrm{pt}}(C_{\mathrm{pt}})$ are both $\geq 0.10$, with 95\% confidence-interval lower bounds above the $0.05$ closure band.
H2 states that $\Delta^{A}_{\mathrm{pt}}(C_{\mathrm{pt}}) < \Delta^{A}_{\mathrm{en}}(C_{\mathrm{pt}})$, supported only if the 95\% confidence interval of the difference excludes zero.

Three pre-registered secondary endpoints replicate the pilot's error analysis on the 448 test texts.
S1 requires that at least half of the undecomposed prompt's false positives on unanimous negatives classify to shortcut bases. This classification, and the groundedness rating in S2, are made by a reading panel of LLM auditors (Claude Fable 5; Anthropic model id \texttt{claude-fable-5}, July 2026). The panel runs two independent readers per text under opposed charitable and skeptical lenses, with a third pass adjudicating disagreements, and never sees the ground-truth labels. S1 and S2 are exploratory analyses since their evidence rests on the panel's judgement rather than on direct and validated measurement.
S2 requires that at least half of the error texts have clause evidence rated at most partially grounded.
S3 requires that the detection clause, the anchor $D$ of Table~\ref{tab:clauses} (id \texttt{D\_entity} in the construct files), fire on fewer than 40\% of texts in both conditions.
Uncertainty is quantified by 95\% bootstrap confidence intervals over 10{,}000 whole-text resamples, fixed seed; mechanics and their interpretation are given in Appendix~D.
Deviations are recorded as dated entries in an append-only log, and all analysis beyond the registered endpoints is defined as exploratory.

\begin{table}[htb]
  \centering
  \small
  \setlength{\tabcolsep}{6pt}
  \begin{tabular}{lrr}
    \toprule
    & \textbf{Pilot} & \textbf{Main} \\
    & (300) & (448) \\
    \midrule
    Pass, round 0 & 268 (89.3\%) & 418 (93.3\%) \\
    Pass, after feedback & 291 (97.0\%) & 442 (98.7\%) \\
    Adjudication & 9 (7 / 2 / 0) & 6 (4 / 2 / 0) \\
    Audit & 60 & 60 \\
    \quad errors corrected & 1 & 1 \\
    Frozen & 300 & 448 \\
    \bottomrule
  \end{tabular}
  \caption{Construction funnel for the $C_{\mathrm{pt}}$ corpus. The main phase runs under the amended entity rule. The audit catch rate beyond the gate is 1.7\% in both phases. Parentheses in the adjudication row give accepted\,/\,edited\,/\,dropped.}
  \label{tab:funnel}
\end{table}

\section{Results}\label{sec:results}
All registered annotations completed.
AMALIA ($P^A_{\mathrm{pt}}$ and $P^A_{\mathrm{en}}$) returned $17{,}952$ responses with no API failures: $16{,}456$ scored annotation calls, $8{,}228$ per instruction condition, and $1{,}496$ calls in the pre-registered robustness pass.
Parse rates reach 100.0\% under both instruction languages, with one unparsable response in the $8{,}228$ Portuguese-instruction calls.
$P^L_{\mathrm{en}}$ (Llama) parsed at 100\%. $P^G_{\mathrm{en}}$ (GPT-OSS) lost 142 of its 8{,}228 calls to runaway reasoning chains (98.3\% parsed). As in the English study, every unrecoverable call, GPT-OSS's 142 and AMALIA's single one, was scored as a negative code at ingest.

The pre-registered robustness pass then removed the JSON constraint from AMALIA's undecomposed prompt.
Unconstrained, the model still returned correctly structured JSON on 746 of 748 texts (99.7\%), in both conditions.
Format discipline is intrinsic to the AMALIA model, not an artefact of constrained decoding.
What follows tests whether that discipline extends to measuring the `authority' construct.

\subsection{Registered endpoints}
\label{sec:results-registered}

On the 448 test texts, AMALIA's recovery gap stays open under both instruction languages.
Under Portuguese instructions, $\Delta^{A}_{\mathrm{pt}} = +0.358$ (95\% CI $[+0.287, +0.434]$).
When AMALIA was used in the instrument with English instructions, $\Delta^{A}_{\mathrm{en}} = +0.436$ (95\% CI $[+0.358, +0.511]$).
In the familiar terms of a standard error, these intervals ask whether the gap would hold on another sample of texts.
Both values exceed the $\Delta \geq 0.10$ band for an open divergence, and both interval lower bounds exceed the $\Delta \leq 0.05$ closure band.
H1 is therefore supported in both conditions.

The full 748-text corpus gives the same descriptive picture.
$F_1^{\mathrm{u}}$ reaches 0.711 under pt-PT instructions and 0.654 under English instructions, while $F_1^{\mathrm{d}}$, by the integration formula, reaches 0.359 and 0.186.
AMALIA agrees with the human annotators almost as well as LLMs eight to thirteen times larger, but recovers the theory-defined construct at half that level under Portuguese instructions and just over a quarter under English ones.

H2 predicted a smaller divergence under Portuguese instructions.
The difference $\Delta^{A}_{\mathrm{en}} - \Delta^{A}_{\mathrm{pt}} = +0.078$, with 95\% CI $[-0.008, +0.163]$.
The pre-registered criterion, an interval excluding zero, was not met, and H2 is therefore not supported under that criterion.
It is worth noting that the lower bound falls below zero by a narrow margin.
The estimate is directionally consistent with the pilot, but the interval is equally compatible with a null effect.
For context, and as exploratory, in the pilot $F_1^{\mathrm{u}}$ was 0.722 under pt-PT instructions against 0.672 under English instructions, and formula recovery gained 0.22 in absolute terms under Portuguese instructions.

The three registered secondary endpoints replicate the pilot's error analysis on the test-set texts. S1 and S2 rest on the LLM reading panel and are exploratory (§\ref{sec:method}); S3 is a direct count.
S1 concerns the basis of the false positives.
Of the 46 unanimous-negative texts that AMALIA's undecomposed prompt coded positive under instrument $P^A_{\mathrm{pt}}$, the reading panel attributed 36 (78\%) to annotation via surface correlates: 14 to generic moral outrage, 10 to the mere presence of an authority figure, 8 to equality look-alikes, and 4 to no identifiable signal.
Ten readings were judged defensible. Inter-reader agreement before adjudication was 63\%.
The 50\% criterion is met (pilot: 88\%).

S2 concerns the evidence.
On 54\% of the error texts, the clause-level evidence segments were rated at most partially grounded, misquoted, or absent from the text in AMALIA's case.
The criterion is met, with a narrower margin than the pilot's 76\%.

S3 concerns detection.
The \texttt{D\_entity} clause fired on 17.9\% of texts under instrument $P^A_{\mathrm{pt}}$ and 12.3\% under $P^A_{\mathrm{en}}$, far below the 40\% bound.
The corpus is not implicated in these errors.
Qualitative re-reads found the foundation present in only 2 of the 46 texts (12 arguable), and 14 of 15 correct-rejection controls were judged to track the construct.
Table~\ref{tab:scorecard} summarizes the registered picture.

\begin{table*}[tb]
  \centering
  \footnotesize
  \setlength{\tabcolsep}{5pt}
  \resizebox{\textwidth}{!}{%
  \begin{tabular}{lllll}
    \toprule
    \textbf{Endpoint} & \textbf{Registered criterion} & \textbf{Estimate} & \textbf{95\% CI} & \textbf{Result} \\
    \midrule
    H1 (PT instructions) & $\Delta \geq 0.10$, CI lower bound $> 0.05$ & $+0.358$ & $[+0.287, +0.434]$ & supported \\
    H1 (EN instructions) & $\Delta \geq 0.10$, CI lower bound $> 0.05$ & $+0.436$ & $[+0.358, +0.511]$ & supported \\
    H2 ($\Delta^{A}_{\mathrm{en}} - \Delta^{A}_{\mathrm{pt}}$) & CI excludes 0 & $+0.078$ & $[-0.008, +0.163]$ & not supported \\
    S1 (shortcut share) & $\geq 50\%$ & 78\% (36/46) & --- & met (exploratory) \\
    S2 (evidence $\leq$ partially grounded) & $\geq 50\%$ & 54\% & --- & met (exploratory) \\
    S3 (\texttt{D\_entity} firing rate) & $< 40\%$ & 17.9\% (pt) / 12.3\% (en) & --- & met \\
    \bottomrule
  \end{tabular}%
  }
  \caption{Pre-registered scorecard, computed on the 448 out-of-sample texts (Zenodo: 10.5281/zenodo.21178967). Intervals are bootstrap 95\% confidence intervals (10{,}000 text resamples, fixed seed).}
  \label{tab:scorecard}
\end{table*}

\subsection{Exploratory analyses}
\label{sec:results-exploratory}

Everything in this subsection is exploratory under the registration.

The reference models answer the question the registered endpoints leave open: whether the open divergence belongs to AMALIA, to the corpus or to the construct.

\begin{figure}[tb]
  \centering
  \includegraphics[width=0.7\linewidth]{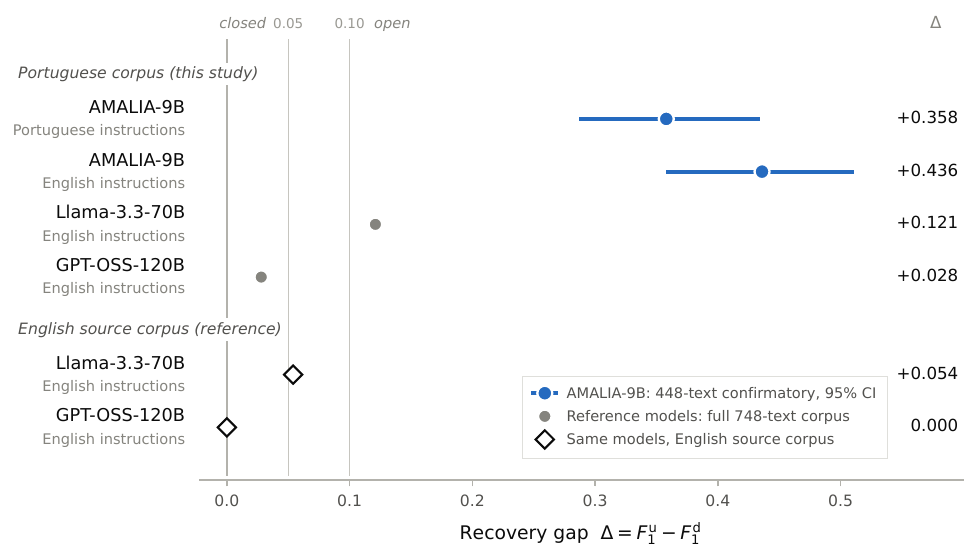}
  \caption{Recovery gap $\Delta = F_1^{\mathrm{u}} - F_1^{\mathrm{d}}$ for each model; larger $\Delta$ means more of the model's agreement with human coders is \emph{not} reproduced when the construct is decomposed into its theory clauses, that is, a larger validity shortfall. AMALIA-9B is plotted at its 448-text confirmatory gaps with 95\% bootstrap confidence intervals (the pre-registered endpoint in Table~\ref{tab:scorecard}); the reference models sit at their full 748-text estimates from Table~\ref{tab:panel}, without intervals. The upper block gives the Portuguese-corpus gaps, with AMALIA-9B under each instruction language and the reference models under English instructions; the lower block gives the same reference models on the English \emph{source} corpus, so the horizontal offset between a model's Portuguese-corpus point and its diamond is the cost of crossing into Portuguese. Vertical lines mark the pre-registered bands: closed ($\Delta < 0.05$) and open ($\Delta \geq 0.10$), with the interval between indeterminate.}
  \label{fig:gaps}
\end{figure}

On the same Portuguese corpus, under the same English instructions, GPT-OSS-120B closes the divergence, with $F_1^{\mathrm{u}}$ 0.753 and $F_1^{\mathrm{d}}$ 0.725, that is, $\Delta^{G}_{\mathrm{en}} = +0.028$, the same closure it shows on the original English corpus.
Llama-3.3-70B does not close it.
Its divergence widens from $+0.054$ on the English corpus to $+0.121$ on the Portuguese one ($F_1^{\mathrm{u}}$ 0.770, $F_1^{\mathrm{d}}$ 0.649), entering the pre-registered open band.
The comparison orders the three annotators by scale: the largest model crosses the language boundary with its closure intact, the mid-sized model widens into the open band, and AMALIA's divergence, $+0.352$ and $+0.468$, sits three to four times above Llama's.
AMALIA's open divergence is therefore primarily a property of the annotator, rather than of the translation, the corpus, or the construct alone; §\ref{sec:discussion} bounds the corpus and language components from these same reference values.
Direct agreement between the models points the same way.
Llama agrees most with the human annotators (0.770), GPT-OSS comes second (0.753), and AMALIA third (0.711).
The 300 pilot texts that shaped the hypotheses sit inside the full 748-text corpus, but AMALIA's recovery gap stays open on the 448 confirmatory texts alone ($\Delta^{A}_{\mathrm{pt}} = 0.358$ and $\Delta^{A}_{\mathrm{en}} = 0.436$, against $0.352$ and $0.468$ on the full corpus). Table~\ref{tab:panel} reports the annotator comparison, and Figure~\ref{fig:gaps} plots the gaps against the pre-registered bands.

The failure profile comes from the pilot condition and is labelled as such.
In the build-up gradient, adding codebook layers pushes AMALIA below its own undecomposed baseline (0.722 to 0.703 under $P^A_{\mathrm{pt}}$; 0.672 to 0.551 under $P^A_{\mathrm{en}}$).
GPT-OSS ends above its baseline (0.753 to 0.778) and Llama ends near its own (0.770 to 0.762), both on the full corpus, where the reference models have no pilot condition.
A logistic regression of the codes produced by AMALIA's own undecomposed prompt on its own answers to the decomposed prompt's clauses placed the equality distinction clause at $\beta \approx 1.65$, as strong a predictor as the true appraisal criteria; 30\% of the undecomposed prompt's positives fired no clause at all.
The two false negatives are real construct misses: parental authority and informal authority, unrecognized in the absence of institutional markers.

\begin{table*}[tb]
  \centering
  \small
  \setlength{\tabcolsep}{5pt}
  \begin{tabular}{llrrrr}
    \toprule
    \textbf{Model} & \textbf{Instructions} & $F_1^{\mathrm{u}}$ & $F_1^{\mathrm{d}}$ & \textbf{Recovery gap} $\Delta$ & \textbf{Parse rate} \\
    \midrule
    \multicolumn{6}{l}{\textit{Portuguese corpus (this study)}} \\
    AMALIA-9B-0626-DPO & Portuguese & 0.711 & 0.359 & $+0.352$ & 100.0\% \\
    AMALIA-9B-0626-DPO & English & 0.654 & 0.186 & $+0.468$ & 100.0\% \\
    Llama-3.3-70B & English & 0.770 & 0.649 & $+0.121$ & 100\% \\
    GPT-OSS-120B & English & 0.753 & 0.725 & $+0.028$ & 98.3\% \\
    \midrule
    \multicolumn{6}{l}{\textit{English source corpus (reference)}} \\
    Llama-3.3-70B & English & 0.697 & 0.643 & $+0.054$ & --- \\
    GPT-OSS-120B & English & 0.737 & 0.737 & $0.000$ & --- \\
    \bottomrule
  \end{tabular}
  \caption{Annotator comparison: undecomposed-prompt $F_1$, integration-formula recovery $F_1$, recovery gap, and parse rate, over the full 748-text corpus. The reference rows are the same models on the English source corpus.}
  \label{tab:panel}
\end{table*}

\section{Discussion and conclusions}\label{sec:discussion}
Two main conclusions follow from these results.
First, as an annotation instrument judged by the standard currently in use, the AMALIA LLM is a reliable annotator.
On the specific test implemented in this paper, annotating the `authority' construct in the MFRC, AMALIA agrees with trained human annotators within less than six points of models eight to thirteen times larger.
Stripped of all output constraints, AMALIA still formats 99.7\% of its responses perfectly.
This capacity to follow output-formatting instructions matters greatly for AMALIA's interoperability with other systems.
Second, as a measurement instrument judged by its capacity to annotate the construct the theory specifies, AMALIA fails in a specific way.
In its best condition, $P^A_{\mathrm{pt}}$, the decomposed prompt recovers a little over half of the agreement the undecomposed prompt achieves.
The other half cannot be attributed to moral foundations theory, but probably to `shortcuts' via surface correlates based, for example, on patterns of vocabulary association.
The error analysis indicates what fills that space.
In 78\% of the unanimous false positives, the reading panel attributes the code to surface bases rather than to the construct: generic moral outrage near a salient authority figure, the mere presence of such a figure, equality look-alikes, or no identifiable signal, without the hierarchical stance and the appraisal of the power-holder that the construct requires.
Crucially, in more than half of the error texts, the evidence AMALIA cited is at best partially grounded in the text.
We also note that the single detection clause fired on fewer than one text in five, in a corpus where half the texts are candidate positives.
Agreement this close to the human annotators and the reference models, built on an annotation process this far from the theory, is therefore precisely the failure mode the framework predicts, demonstrated for the first time on a national model in its own language.
It should also be noted that the study's scope is limited to one construct and one corpus.

The exploratory comparison turns this limitation into a deeper diagnosis.
If the failure were explained by the Portuguese transcreated corpus, every LLM annotation instrument used would have failed too.
GPT-OSS-120B closes the divergence on this corpus to $+0.028$, the same closure it shows on the original English corpus.
The instructions do not explain the failure either.
GPT-OSS annotated with the same English instructions used with AMALIA and, even so, closed the divergence that AMALIA keeps open in both instruction languages.
What remains is the LLM annotator.
As a rough decomposition rather than a formal estimate, GPT-OSS's residue on this corpus suggests that a corpus artefact large enough to explain AMALIA's gap is unlikely; the widening of Llama's divergence from $0.054$ to $0.121$ places the language effect at about $0.07$; and close to $0.25$ of AMALIA's $0.352$ remains.

A construct calibrated to detection grade on a 120-billion-parameter model exceeds what a nine-billion-parameter model can execute clause by clause.
As §\ref{sec:background} sets out for the recovery gap, the executable grain of a construct belongs to the construct--model pair, not the construct alone \citep{pita:2026a}.
The practical consequence is not that small national models, like AMALIA, are useless as annotation instruments; it is that the calibration does not transfer.
In principle, the instrument can be recalibrated at a coarser grain, matched to what a 9B model reliably detects.
But, for validation, the grain of a construct is not a free parameter: it has a lower bound that does not belong to the annotator, it belongs to the theory.
Below that bound, the clauses cease to correspond to the conditions the theory states and become miniature holistic judgements, more specific than the undecomposed prompt but equally opaque about the process that produces them.

The data suggest that this transferred calibration places AMALIA below that bound.
The evidence is twofold.
On one hand, the instrument's most superficial clause, the mere detection of the authority entity, fires on fewer than one fifth of the texts.
On the other, the procedural readings, the structured re-reading of the unanimous errors that classifies the basis of each judgement and the grounding of the cited excerpts, show that the holistic agreement rests on surface shortcuts.
A recalibration at that coarser grain would produce an annotator reliable in behaviour but not valid in construct.
In Messick's terms (§\ref{sec:background}), it would lack the substantive aspect of construct validity.
This makes AMALIA a useful tool for triage, pre-annotation and monitoring with downstream human review; it is not enough for measurement that supports inference about the construct.
Establishing whether this bound is one of principle or of instruction would require running the calibration loop on the 9B model itself, widening the grain to the construct's minimal articulation: the experiment this study motivates but does not carry out.
What this study rules out is the shortcut the field has been taking, that of assuming that agreement certifies the instrument that produced it, whatever it is. Agreement is attainable below the grain that articulates the construct, and that is exactly where AMALIA attains it.

The question of the home-language advantage splits into two theses that these data treat differently, because language enters them at different points of the instrument.
In one, it is the language of the annotated text; in the other, the language of the instructions.
The agreement thesis, that a national model should agree best with human annotators on text in its own language, is refuted descriptively.
On the Portuguese corpus, the two multilingual reference models agree more than AMALIA (0.770 and 0.753 against 0.711, on the holistic measure); there is no home-turf advantage.
The validity thesis, that Portuguese instructions should narrow AMALIA's recovery gap, remains open.
The pre-registered criterion was not met.
The observed direction was the predicted one, but the interval ($+0.078$, 95\% CI $[-0.008, +0.163]$) is equally compatible with the absence of an effect.
The strong signal is exactly where the theory would place it, at the grain of the clauses, where the judgement depends on following the instruction and not on surface shortcuts; formula recovery gained 0.17 in $F_1$ under Portuguese instructions on the full corpus (0.22 in the pilot).
Resolving the validity thesis requires a corpus sized for an effect on the order of $+0.08$, which points to a thousand texts; once that corpus is built, auditing each new version of the model costs about two dollars of compute, and the test becomes routinely repeatable at each checkpoint.
What no one should do with these data is fuse the two theses into a slogan, whether for or against national models.

The wider implication is for national-model programmes, not for this model in particular.
The sovereignty case for these models is at bottom an auditability case (§\ref{sec:intro}), and AMALIA honours it unusually fully, with open weights, data and code under the Apache 2.0 licence \citep{simplicio:2026a}; that openness is what made this study possible, at about thirty euros and ninety minutes of expert reading.
Validity evaluation in the sense of \citet{messick1995validity}, however, is not yet part of model assessment: current practice, AMALIA's report included, rests on knowledge and fluency benchmarks scored by an automatic judge (§\ref{sec:background}) \citep{simplicio:2026a}.
The team built native benchmarks because translation loses what matters in the local variety, which is itself a validity argument; carried to its conclusion, it demands validity evidence not only for the benchmarks a model is scored on but for the uses a community will put it to, measurement chief among them.
A battery of recovery-gap analyses over a set of constructs is the audit the sovereignty argument promises and the evidence the validity argument requires; it fits any programme's budget and belongs in future versions of the model.

These findings separate three kinds of trust that discussions of sovereign models tend to fuse.
\textit{Operational trust} concerns conduct: the model follows instructions, produces parseable output, and can be held and inspected. AMALIA earns it in full.
\textit{Performance trust} concerns agreement: the model's codes match those of trained annotators often enough to be useful. AMALIA earns much of it, within six points of far larger models.
\textit{Epistemic trust} concerns warrant: the codes are produced by the inferences the construct's theory specifies, so that measurements taken with the instrument support conclusions about the construct. This is the trust a scientific use requires, and it is the one AMALIA, on this construct, does not yet earn.
Sovereignty buys the conditions for the first two, and the auditability needed to test the third; it cannot buy the third itself.
The distinction hands programmes a vocabulary for release notes and procurement alike: which trust a benchmark certifies, and which it leaves open.

The transcreation apparatus is the part of this study that any other language community can adopt without adopting our conclusions.
It combines transcreation anchored in the register, length and referents of the original, and blind to the annotated codes.
This design creates a hermetic separation between those who build the corpus and the models evaluated, an automatic five-criterion gate that errs by excess of rigour, human adjudication that outranks the gate, and a back-translation audit that catches what the models' fluency may `hide'.
Together, these elements processed 748 ground-truthed texts across a language boundary, with six human edits, full provenance for each text, and a duly justified rather than assumed code transfer.
The verification numbers are the reusable result: 97\% and 99\% pass rates on the automatic gate across the two phases, and a stable 1.7\% of errors that only a native reader catches.
Any calibrated instrument and its ground truth can cross a language boundary this way, at negligible cost, before anyone trusts an annotator on the far side.

The study has scoping limitations that create space for future work.
The ground-truth codes are inherited from the English originals; verification defends the transfer, but a residual risk of code drift remains, which is why the unanimous stratum, the least exposed to that drift, anchors the error analysis.
What verification defends is the preservation of meaning and stance; cross-linguistic equivalence of the construct itself is not established by this design, and §\ref{sec:background} places it outside the recovery gap's scope.
The Portuguese construct is a faithful rendering, not a recalibration.
This was the aim, since portability was under test, but it also means these results do not say how well a construct calibrated \emph{for} AMALIA would perform.
The study covers one construct, one national model and one corpus register (Reddit posts); the shortcut mechanism identified here may be another in other constructs, other models or other registers.
Because there is no base checkpoint against which to compare, the failure profile cannot be attributed to pretraining or to post-training, including the obvious candidate this design cannot isolate: the largely synthetic Portuguese instruction data, generated by Gemma (§\ref{sec:background}).
The hypotheses were informed by the pilot, as the pre-registration declares, and all confirmatory tests ran out of sample.
The test of H2 was underpowered; the hypothesis was not refuted. The pilot-condition diagnostics in §\ref{sec:results-exploratory} are exploratory, and do not constitute confirmed results.

The S1 and S2 readings rest on an LLM panel, not on human raters; the panel never sees the ground-truth labels, and restricting S1 to the texts where the two opposed lenses agreed without adjudication raises the shortcut share from 78\% to 88\%, so the finding does not depend on the judge.
The full protocol, its robustness analysis, and the shipped verdicts are given in Appendix~E.

AMALIA shares its name with the country's most celebrated voice, and in the days after release, the question asked everywhere was the natural one: what can the AMALIA LLM do?
This paper analyses a narrow version of the question, with a fixed measurement standard and a public protocol.
The answer has two parts.
AMALIA handles these European-Portuguese texts fluently, meets the requested format almost without a slip, and, asked whether a text honours or defies authority, answers, with a frequency remarkable for its size, as the human annotators do.
But asked to show its work, to find the authority, establish the stance and rule out the look-alikes, the pieces do not assemble in its answers.
AMALIA hears moral outrage near someone powerful and calls it authority.
In the errors the panel read closely, half the evidence AMALIA cites is at best partially grounded in the text. A larger model, given the same Portuguese texts, showed its work and matched its own answers; AMALIA could not, in either instruction language. So is AMALIA a valid instrument for coding authority? An instrument is valid for a construct only as far as it can execute the theory that defines it. By that measure, not yet, and not at this grain, though its codes read as if it were; and that is exactly the result.

The lesson does not stay in Portugal. Every language community that receives a national model will be tempted to hand it the work of measurement, whether annotating what its citizens say, tracking what its media value, or classifying what its institutions produce. Before that, the community should ask the model not whether it agrees with the human annotators, but whether it can show the work that agreement is supposedly made of. The test is cheap, this one cost less than a launch's celebration dinner, and the instrument to run it now exists in Portuguese. Agreement was the wrong question to ask of AMALIA. It is the wrong question to ask of any of them.

\section*{Acknowledgements}

The author thanks the AMALIA team for releasing the model with open weights, data and code, and the MFRC authors for confirming the deduplication artefact reported in \S\ref{sec:method}.

\section*{Ethical considerations}

This study analysed secondary, publicly available text (the Moral Foundations Reddit Corpus) and did not involve human participants, patients, or animals. Ethical approval was therefore not required.

\section*{Consent to participate}

Not applicable; the study did not involve human participants.

\section*{Consent for publication}

Not applicable.

\section*{Funding}

This work was produced at the Artificial Intelligence, Social Interaction and Complexity Laboratory, supported by the CICANT research unit at Universidade Lus\'ofona (UID/05260/2025, \url{https://doi.org/10.54499/UID/05260/2025}).

\section*{Declaration of conflicting interests}

The author declares no competing interests.

\section*{Use of generative artificial intelligence}

Claude (Anthropic, \texttt{claude-fable-5}, July 2026) served as the S1/S2 reading panel under the pre-registered protocol of \S\ref{sec:method}. It also assisted with analysis tooling and manuscript preparation, under the author's method design and specific instructions; the author double-checked all of this assistance, and analytical decisions, the verification of every reported number against the registered artifacts, and the final text are the author's responsibility.

\section*{Data and code availability}

The replication package---the paired English/European-Portuguese corpus with ground truth, the frozen English and Portuguese constructs, the per-annotator annotation and evidence tables, and the analysis scripts---is openly available at \url{https://github.com/sejkko/amalia-llm-on-authority} and archived at Zenodo (DOI:~\href{https://doi.org/10.5281/zenodo.21275660}{10.5281/zenodo.21275660}). Every pre-registered endpoint can be independently replicated from it. The pre-registration is deposited separately (DOI:~\href{https://doi.org/10.5281/zenodo.21178967}{10.5281/zenodo.21178967}). Source texts and ground-truth labels derive from the Moral Foundations Reddit Corpus \citep{trager2022moral}, released under CC-BY~4.0.

\appendix

\section{Verification gate and adjudication}

The verification gate applies five criteria.
It requires each generated text to preserve meaning and stance, use the pt-PT variant rather than pt-BR or a mix, sound natural, and keep entities stable.
Failures regenerate with the verifier's feedback, up to three attempts.
Texts that exhaust their attempts go to human adjudication, whose decision is final and outranks the machine verifier.
Table~\ref{tab:funnel} reports the funnel.
In the pilot phase, 89.3\% of texts passed at the first round and 97.0\% after feedback rounds; the remaining nine went to human adjudication.
Of the nine adjudicated texts, seven were machine-verifier false positives on genuine European-Portuguese idiom, and two were edited.
In the main phase, the amended entity rule is explicit in the funnel, with 93.3\% passing at the first round, 98.7\% after feedback, and six adjudications, four accepted and two edited, none dropped.
The verification gate errs by excess of rigour, by design, and the human layer corrects it.

Pipeline sampling parameters: The transcreation pipeline operates with DeepSeek-V4-Pro to generate pt-PT texts at temperature 0.7, Qwen3-235B verifies at temperature 0, and DeepSeek-V4-Pro back-translates at temperature 0.

\section{Back-translation audits}

The verification gate reads fluency, so it can pass a fluent text that quietly fails to capture meaning.
Two stratified audits, one per phase, each covered a sample of sixty texts, drawn at random in each phase with seed 20260703.
The registrant, a native speaker of European Portuguese, read these back-translations against the original English sources.
Each audit, one per phase, pilot and main, caught exactly one error the earlier gate had passed, a meaning failure in the pilot and a suppressed opening sentence in the main phase, a catch rate of 1.7\% beyond the gate.
Both errors were corrected as logged human edits.

\section{Serving and output configuration}

Output limits: 512 tokens for the instructed models and 4K for GPT-OSS. Logged fallback tiers apply to responses that resist structured parsing.

\section{Bootstrap mechanics}

Uncertainty is quantified by 95\% bootstrap confidence intervals, computed by the percentile method over 10{,}000 resamples that draw whole texts with replacement, so a text's clause and holistic codes stay together (seed 20260703).
Because the coding is deterministic and the ground-truth labels are fixed, these intervals reflect sampling across texts, not the reliability of the coding itself; a lower bound above the closure band means the open gap survives resampling.

\section{Reading-panel protocol and robustness}

The S1 and S2 readings rest on an LLM panel, not on human raters.
The two readers and the judge are instances of one model prompted under opposed lenses, so the 63\% pre-adjudication agreement measures prompt-induced divergence, not the independence of two human minds.
Four features bound the risk.
The readers never see the ground-truth labels.
Fourteen of the fifteen correct-rejection controls, read blind alongside the errors, come back judged as rejections for construct reasons.
Restricting S1 to the twenty-five error texts on which the two lenses agreed without adjudication raises the shortcut share from 78\% to 88\%, so the finding does not depend on the judge.
And the registered criterion leaves S1 a wide margin, the estimate surviving up to thirteen of its thirty-six shortcut attributions being overturned, while S2 clears its criterion by two texts and should be read as confirmed but fragile.
The full targets, verdicts, and reading protocol ship with the release, so the reading can be repeated by any panel, human or machine.

\end{document}